\documentclass{bvm}

\begin{document}

\newcommand{\bvmyear}{2026}

\selectlanguage{english}

\title{EvalBlocks}
\subtitle{A Modular Pipeline for Rapidly Evaluating Foundation Models in Medical Imaging}
\titlerunning{EvalBlocks}

\author{
	\fname{Jan} \lname[0009-0007-0047-9225]{Tagscherer} \affiliation{Radboud University Medical Center} \authorsEmail{jan.tagscherer@radboudumc.nl} \street{Geert Grooteplein Zuid} \housenumber{10} \zipcode{6525 GA} \city{Nijmegen} \country{The Netherlands} \isResponsibleAuthor,
	\fname{Sarah} \lname{de~Boer} \affiliation{Radboud University Medical Center} \authorsEmail{sarah.deboer@radboudumc.nl} \street{Geert Grooteplein Zuid} \housenumber{10} \zipcode{6525 GA} \city{Nijmegen} \country{The Netherlands},
	\fname{Lena} \lname{Philipp} \affiliation{Radboud University Medical Center} \authorsEmail{lena.philipp@radboudumc.nl} \street{Geert Grooteplein Zuid} \housenumber{10} \zipcode{6525 GA} \city{Nijmegen} \country{The Netherlands},
	\fname{Fennie} \lname{van~der~Graaf} \affiliation{Radboud University Medical Center} \authorsEmail{fennie.vandergraaf@radboudumc.nl} \street{Geert Grooteplein Zuid} \housenumber{10} \zipcode{6525 GA} \city{Nijmegen} \country{The Netherlands},
	\fname{Dré} \lname{Peeters} \affiliation{Radboud University Medical Center} \authorsEmail{dre.peeters@radboudumc.nl} \street{Geert Grooteplein Zuid} \housenumber{10} \zipcode{6525 GA} \city{Nijmegen} \country{The Netherlands},
	\fname{Joeran} \lname{Bosma} \affiliation{Radboud University Medical Center} \authorsEmail{joeran.bosma@radboudumc.nl} \street{Geert Grooteplein Zuid} \housenumber{10} \zipcode{6525 GA} \city{Nijmegen} \country{The Netherlands},
	\fname{Lars} \lname{Leijten} \affiliation{Radboud University Medical Center} \authorsEmail{lars.leijten@radboudumc.nl} \street{Geert Grooteplein Zuid} \housenumber{10} \zipcode{6525 GA} \city{Nijmegen} \country{The Netherlands},
	\fname{Bogdan} \lname{Obreja} \affiliation{Radboud University Medical Center} \authorsEmail{bogdan.obreja@radboudumc.nl} \street{Geert Grooteplein Zuid} \housenumber{10} \zipcode{6525 GA} \city{Nijmegen} \country{The Netherlands},
	\fname{Ewoud} \lname{Smit} \affiliation{Radboud University Medical Center} \authorsEmail{ewoud.smit@radboudumc.nl} \street{Geert Grooteplein Zuid} \housenumber{10} \zipcode{6525 GA} \city{Nijmegen} \country{The Netherlands},
	\fname{Alessa} \lname{Hering} \affiliation{Radboud University Medical Center} \authorsEmail{alessa.hering@radboudumc.nl} \street{Geert Grooteplein Zuid} \housenumber{10} \zipcode{6525 GA} \city{Nijmegen} \country{The Netherlands}
}

\authorrunning{Tagscherer et al.}

\institute{
Diagnostic Image Analysis Group, Radboud University Medical Center, Nijmegen, The~Netherlands
}

\email{jan.tagscherer@radboudumc.nl}

\maketitle

\begin{abstract}
	Developing foundation models in medical imaging requires continuous monitoring of downstream performance. Researchers are burdened with tracking numerous experiments, design choices, and their effects on performance, often relying on ad-hoc, manual workflows that are inherently slow and error-prone. We introduce \emph{EvalBlocks}, a modular, plug-and-play framework for efficient evaluation of foundation models during development. Built on Snakemake, EvalBlocks supports seamless integration of new datasets, foundation models, aggregation methods, and evaluation strategies. All experiments and results are tracked centrally and are reproducible with a single command, while efficient caching and parallel execution enable scalable use on shared compute infrastructure. Demonstrated on five state-of-the-art foundation models and three medical imaging classification tasks, EvalBlocks streamlines model evaluation, enabling researchers to iterate faster and focus on model innovation rather than evaluation logistics. The framework is released as open source software at \url{https://github.com/DIAGNijmegen/eval-blocks}.
\end{abstract}

\section{Introduction}

Foundation models have shown great promise in medical imaging, learning semantically rich embeddings from large-scale pretraining that can then be used for few-shot adaptation to data-scarce tasks. When integrated into downstream pipelines, these pretrained models can substantially accelerate development and improve performance across diverse clinical applications. While this quality is appealing, developing these models involves a multitude of design choices, such as data sampling, architecture selection, and training strategy. This results in an iterative development process during which it is important to continuously estimate a model's downstream performance and gain insights into the impact of training and architecture choices.

The evaluation of foundation models is often performed using bespoke scripts while also managing compute resources and organizing experiments. This unnecessary difficulty slows iteration, complicates reproducibility, and shifts focus away from improving models themselves.

The emergence of foundation models has prompted the creation of benchmarks to compare their downstream performance on various medical imaging tasks. Wang et al.~\cite{3865-02} define clinically relevant tasks for a systematic comparison, Jin et al.~\cite{3865-03} assess fairness across datasets, tasks, and sensitive attributes, and the UNICORN challenge~\cite{3865-01} evaluates submitted models on multimodal tasks. While valuable for standardized comparison, these benchmarks focus on comprehensiveness over rapid evaluation during model development.

Similar needs for lightweight evaluation have been addressed in other domains. Hugging Face's LightEval~\cite{3865-04} supports the rapid assessment of large language models, and NVIDIA's NeMo Evaluator SDK~\cite{3865-15} aims to make LLM evaluation robust, reproducible, and scalable.

In medical imaging, however, a comparable tool for efficient and reproducible model evaluation is lacking. We address this gap with \emph{EvalBlocks}, a modular, extensible, and cluster-ready pipeline based on Snakemake~\cite{3865-05} and designed for efficient, reproducible assessment of foundation models in medical imaging. We demonstrate the utility of our pipeline by evaluating five recent foundation models across three malignancy classification tasks.

In summary, our contributions include:
\begin{itemize}
	\item A modular, extensible, and efficient evaluation framework for foundation models in medical imaging that is available as open source software.
	\item A demonstration of the pipeline that evaluates five foundation models on three medical imaging classification tasks.
\end{itemize}

\section{Materials and methods}

\subsection{Architecture overview}

\begin{figure}[b]
	\includegraphics[width=\linewidth]{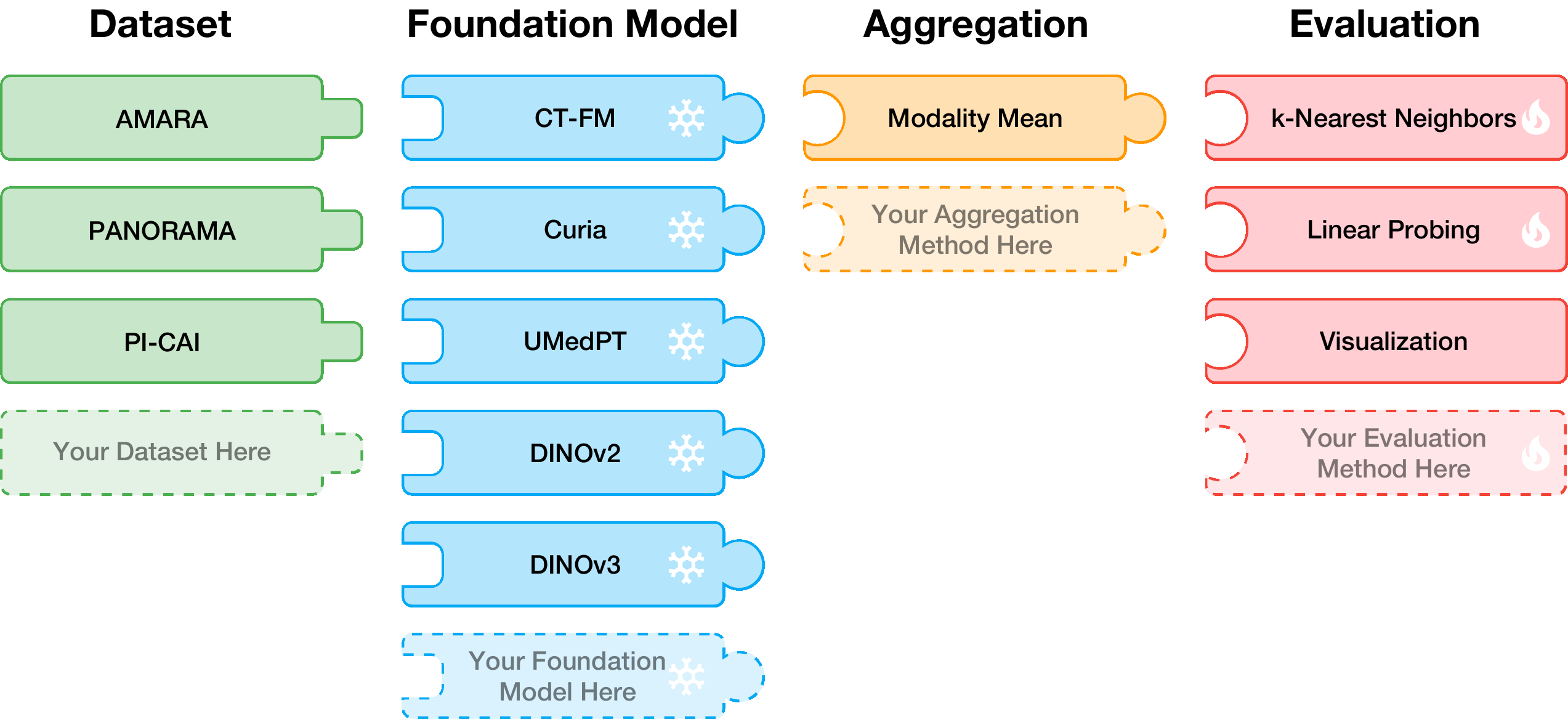}
	\caption{In our framework, pipeline steps are implemented as self-contained blocks. Foundation models embed input patches, and these feature embeddings can be optionally aggregated and then evaluated. The pipeline blocks can be freely extended and plugged into each other, enabling fast, reproducible, and customizable evaluation during foundation model development.}
	\label{3865-pipeline}
	\altText{An illustration that frames the pipeline as a set of puzzle pieces of datasets, foundation models, aggregation methods, and evaluation methods that can be freely plugged into each other.}
\end{figure}

Figure \ref{3865-pipeline} illustrates the pipeline, composed of independent Snakemake rules that define their input-output dependencies and resource requirements. They are automatically executed when their required inputs are available. Rules are grouped into three categories: (1) Feature models that transform input patches into embeddings, (2) optional aggregation steps, and (3) evaluation procedures. Intermediate outputs are cached for efficient reuse.

Experiments are recorded declaratively in a configuration file, specifying datasets, models, and evaluation methods. The pipeline can run selected experiments or all configured combinations on demand and supports distributed execution in cluster environments such as Slurm~\cite{3865-06}, running computational steps in parallel wherever possible.

We demonstrate the framework's utility by implementing a set of blocks that allow for the evaluation of five foundation models across three medical imaging classification tasks. The goal of these experiments is not to advance state-of-the-art performance, but to demonstrate how EvalBlocks accelerates experimental iteration.

\subsection{Datasets}

We evaluate on three patch-level malignancy classification tasks derived from the AMARA (in-house), PANORAMA~\cite{3865-07}, and PI-CAI~\cite{3865-08} datasets. Each dataset provides training and test splits across five folds. For all datasets, we extract patches of size $224 \times 224 \times 16$ along with malignancy labels. Input data is preprocessed according to the specifications provided by each model's authors. For models that can handle three-dimensional input data, we use the entire patch. For two-dimensional architectures, we input the central slice. Finally, DINOv2~\cite{3865-12} and DINOv3~\cite{3865-13} have been trained on natural images. For these models, we interpret the input slices as grayscale images with values between $0$ and $255$.

From the AMARA dataset's CT scans, we extract 161 malignant and 502 benign pulmonary nodules from 320 patients with ground-truth labels determined by pathological confirmation.

The PANORAMA dataset~\cite{3865-07} yields 675 CT patches of healthy pancreatic tissue and 675 patches with ductal adenocarcinoma.

Finally, we produce 219 MR patches depicting prostate carcinoma and 219 patches with healthy prostate tissue from the public test set of the PI-CAI challenge~\cite{3865-08}.

\subsection{Foundation models}

We evaluate five foundation models: CT-FM~\cite{3865-09} is the only model that has been trained on three-dimensional CT scans as its only modality, while the remaining medically-focused models process two-dimensional but multi-modal input data. Curia~\cite{3865-10} has been created through unsupervised training on a large dataset of medical images. UMedPT~\cite{3865-11} is the only model in our evaluation that has been trained in a supervised manner. Finally, we also include DINOv2~\cite{3865-12} and DINOv3~\cite{3865-13}, which have been trained on natural images rather than medical imaging data. The public release includes preconfigured blocks for these models, including all necessary preprocessing steps, enabling immediate plug-and-play use.

\subsection{Aggregation methods}

For demonstration purposes, we aggregate the embeddings of our MRI dataset by computing the element-wise mean of feature vectors across modalities to assess whether combining complementary contrasts improves downstream performance.

Beyond this example, the framework supports defining custom aggregation modules, enabling more complex strategies such as weighted averaging, attention-based fusion, or case-level pooling.

\subsection{Evaluation strategies}

We implement three interchangeable evaluation strategies that operate on the optionally aggregated feature embeddings.

First, we fit a k-Nearest Neighbors classifier with $k \in \{10, 20, 100, 200\}$ on the training features and report accuracy and AUC on the test split; results for $k = 20$ are shown in the following. Second, we train a single linear layer using cross entropy loss with a learning rate of $1e-5$ and evaluate its accuracy and AUC. Third, we generate visual analyses by applying linear discriminant analysis, principal component analysis, and t-SNE to the feature embeddings, providing interpretable plots of the learned representations.

\section{Results}

We evaluated all combinations of foundation models, aggregation methods, and evaluation strategies using the EvalBlocks pipeline. This produced a comprehensive set of metrics and visualizations for each configuration, demonstrating that the pipeline executes and records experiments in an automated and reproducible manner.

\begin{figure}[b]
	\includegraphics[width=\linewidth]{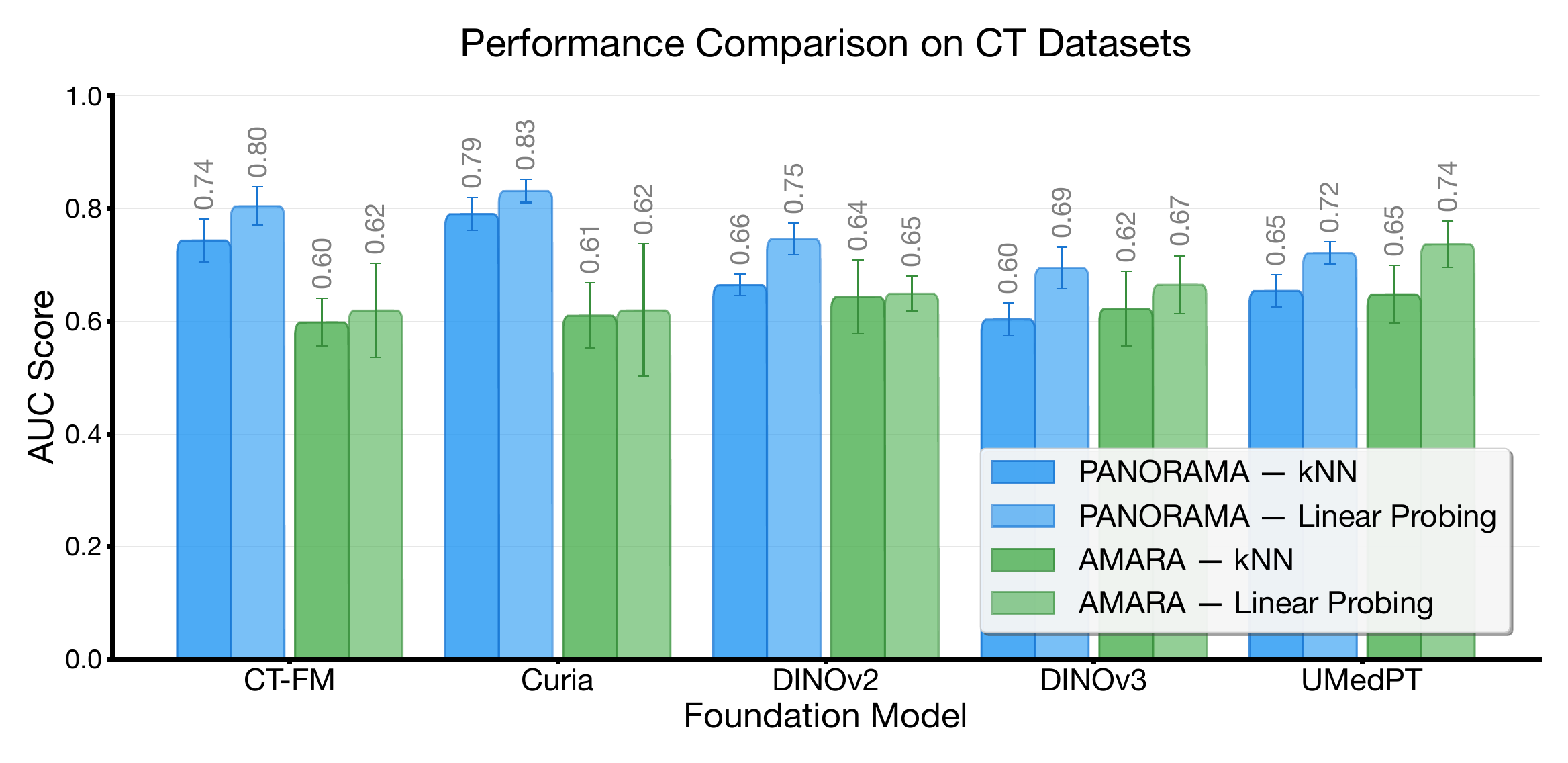}
	\caption{A visualization of model results on our CT datasets created by running EvalBlocks, with error bars depicting the standard deviation across folds. While CT-FM~\cite{3865-09} and Curia~\cite{3865-10} perform best on PANORAMA~\cite{3865-07}, UMedPT~\cite{3865-11} is slightly more accurate on AMARA. Our pipeline allows for fast and automated comparison between models and checkpoints.}
	\label{3865-ct-results}
	\altText{A bar chart showing the five foundation models and their kNN and linear probing performance on the PANORAMA and AMARA datasets.}
\end{figure}

Fig.~\ref{3865-ct-results} depicts model performance on our two CT datasets, showcasing how EvalBlocks can be used to estimate the difficulty of a downstream task and compare models.

\begin{figure}[b]
	\includegraphics[width=\linewidth]{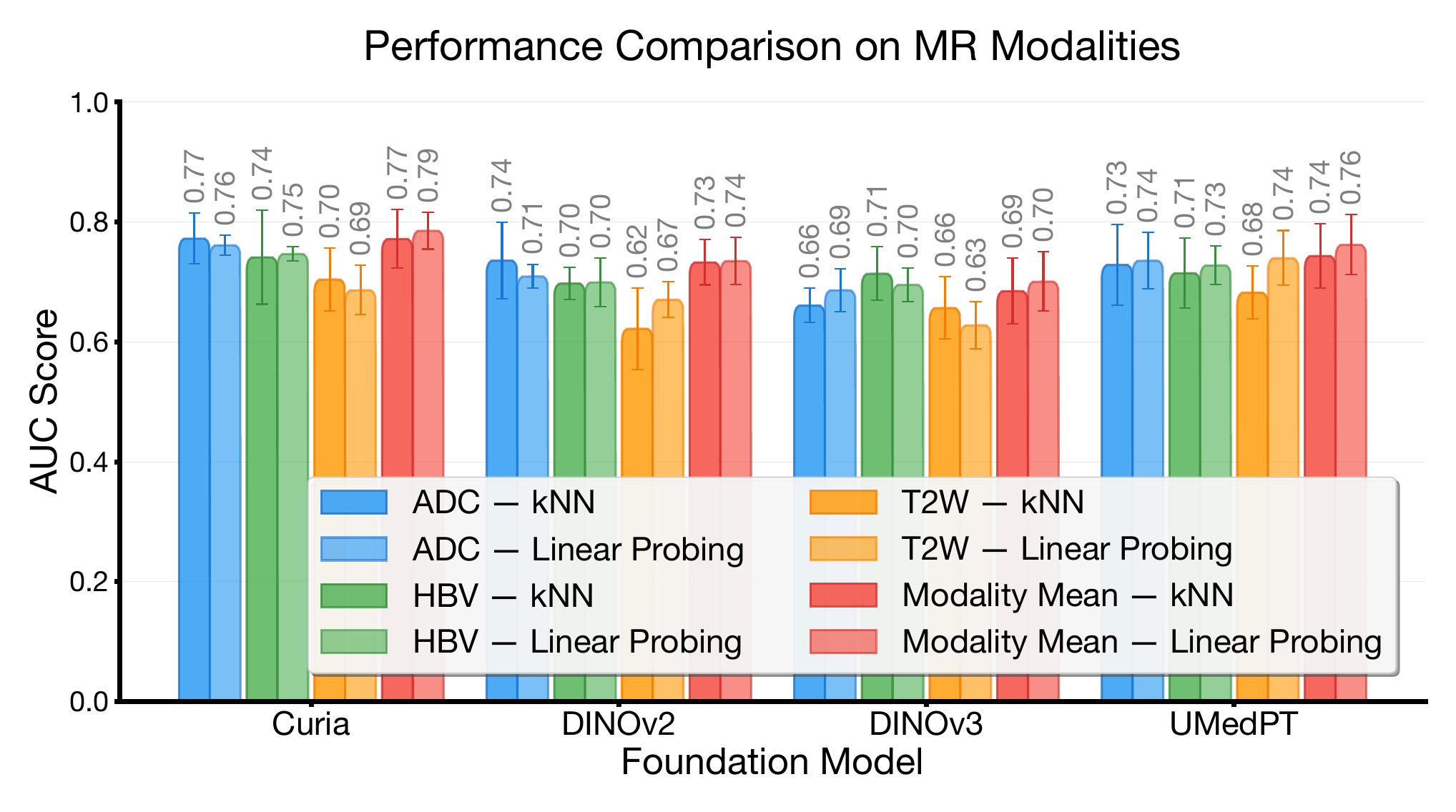}
	\caption{EvalBlocks also enables evaluation across modalities and aggregation methods, here for the PI-CAI~\cite{3865-08} dataset. Error bars denote the standard deviation across folds. Overall, ADC is the most informative modality for malignancy discrimination. The modality mean aggregation emerges as a well-performing strategy for this task. Our framework enables researchers to easily prototype aggregation methods.}
	\label{3865-mr-results}
	\altText{A bar chart showing four foundation models and their kNN and linear probing performance on the PI-CAI dataset. For each model, its performance is stratified across modalities and aggregation methods.}
\end{figure}

Fig.~\ref{3865-mr-results} focuses on our framework's ability to evaluate across modalities and aggregation methods, allowing for fast prototyping of the latter and informed selection of inputs.

\begin{figure}[b]
	\centering
	\setlength{\figwidth}{0.49\textwidth}
	\begin{subfigure}{0.7\textwidth}
		\includegraphics[width=\textwidth]{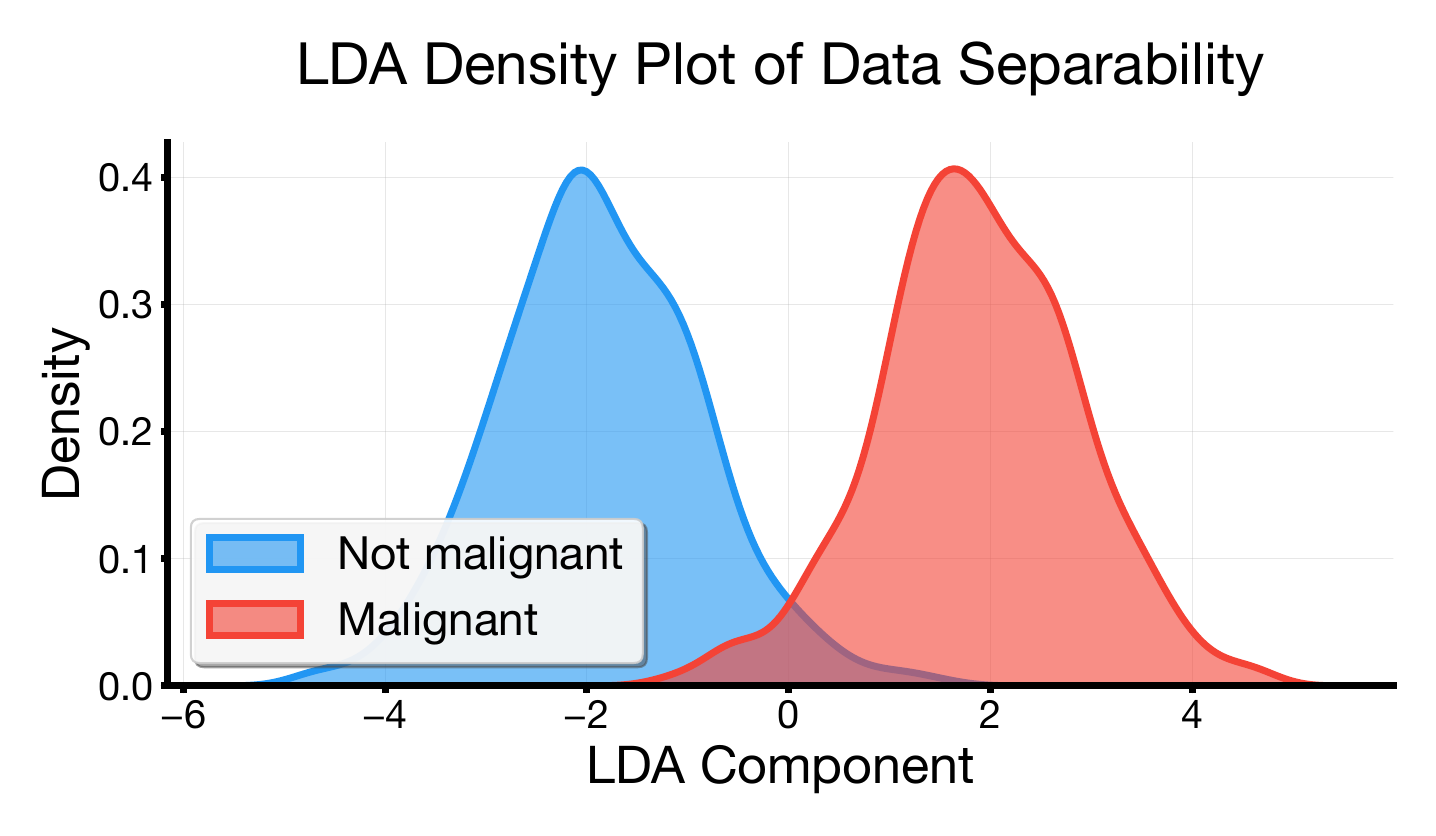}
		\caption{Linear Discriminant Analysis.}
		\label{3865-visualisation-results-lda}
		\altText{An LDA density plot of the features, showing two distinct peaks.}
	\end{subfigure}
	\hfill
	\begin{subfigure}{\figwidth}
		\includegraphics[width=\textwidth]{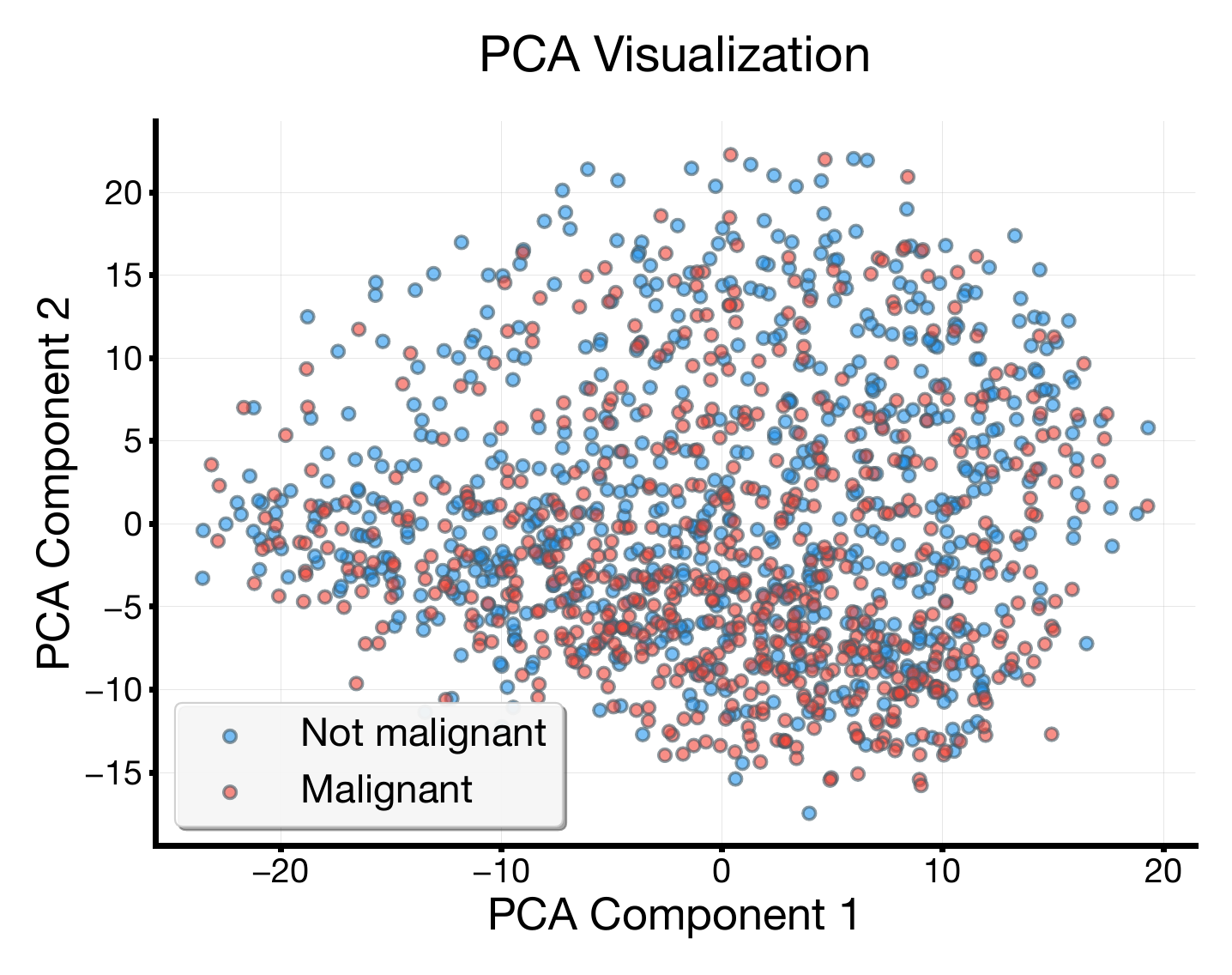}
		\caption{Principal Component Analysis.}
		\label{3865-visualisation-results-pca}
		\altText{A PCA scatterplot of the features without clear clusters.}
	\end{subfigure}
	\hfill
	\begin{subfigure}{\figwidth}
		\includegraphics[width=\textwidth]{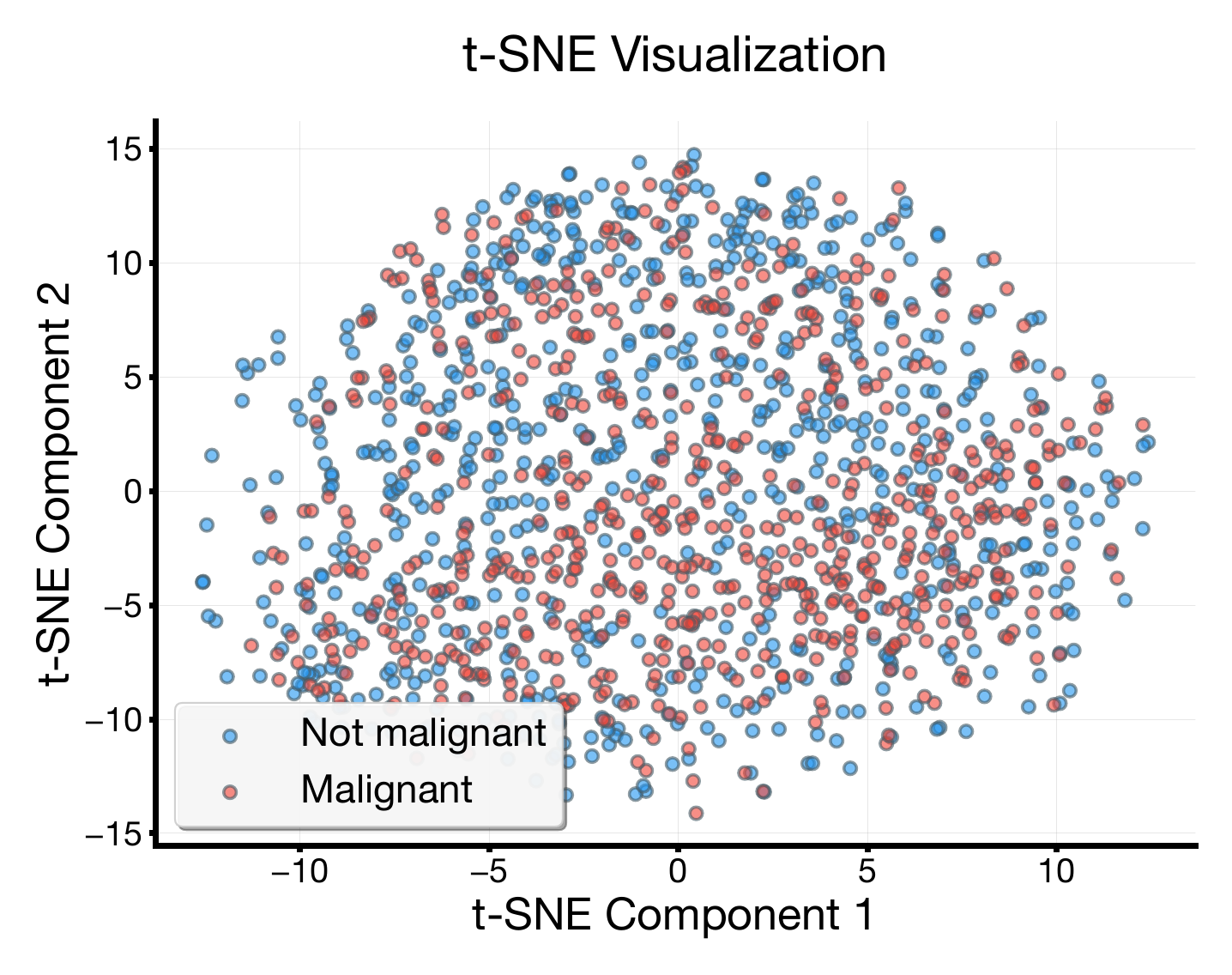}
		\caption{t-Distr. Stochastic Neighbor Embedding.}
		\label{3865-visualisation-results-tsne}
		\altText{A t-SNE scatterplot of the features without clear clusters.}
	\end{subfigure}

	\caption{Visualizations of the feature embeddings of Curia~\cite{3865-10} on the first fold of the PANORAMA dataset~\cite{3865-07}. While PCA and t-SNE yield no clusters, LDA shows two distinct peaks for the two classes. This reveals that the model produces linearly separable feature embeddings for this task in label-dependent directions, but not in directions of maximum variance or local neighborhood structures. EvalBlocks produces these visualizations for all folds, datasets, and models, allowing deeper analysis where necessary.}
	\label{3865-visualisation-results}
	\altText{LDA, PCA, and t-SNE plots of patch features.}
\end{figure}

Finally, Fig.~\ref{3865-visualisation-results} showcases how the visualization block can enable a more thorough analysis of feature embeddings produced by the foundation models.

During the evaluation of these models, caching avoided recomputing embeddings across experiments. In combination with the framework's parallel execution capabilities, this reduced wall-time substantially.

\section{Discussion}

EvalBlocks provides a modular and efficient framework for evaluating foundation models in medical imaging. In this study, we demonstrated its flexibility and utility by evaluating five foundation models across three downstream classification tasks with minimal configuration effort. The modular design facilitates the rapid integration of datasets, models, aggregation strategies, and evaluation methods. The framework's efficient caching and centralized experiment tracking substantially reduces both computational and manual effort. Furthermore, EvalBlocks can run locally, which made the assessment of foundation models on in-house datasets possible.

We note that, as the number of datasets and models grows, the combinatorial space of possible evaluations expands quickly. EvalBlocks mitigates this by leveraging Snakemake's caching and parallelization capabilities and by allowing users to selectively run subsets of experiments.

While existing benchmarks are useful as static leaderboards for foundation models, they are not suited for iterative model development. EvalBlocks fills this gap by enabling reproducible, transparent, and scalable evaluation during model development, thus bridging the gap between large-scale benchmarking and practical experimentation.

Future work will expand EvalBlocks to additional task types such as segmentation and detection. Integrating the framework with existing popular platforms like Hugging Face will allow for better community collaboration. By reducing the burden of evaluation logistics, EvalBlocks allows researchers to focus on improving model architectures, training strategies, and downstream adaptation.

\section*{Note}
\small This is a preprint of the following work: Tagscherer et al., ``EvalBlocks: A Modular Pipeline for Rapidly Evaluating Foundation Models in Medical Imaging,'' published in \textit{Bildverarbeitung f{\"u}r die Medizin 2026: Proceedings of the German Conference on Medical Image Computing, March 15--17, 2026, L{\"u}beck}, edited by Heinz Handels, Katharina Breiniger, Thomas Deserno, Andreas Maier, Klaus Maier-Hein, Christoph Palm, and Thomas Tolxdorff, Springer Vieweg, Wiesbaden, 2026. This version may differ from the final published version. The final authenticated version is available online at: \url{https://doi.org/10.1007/978-3-658-51100-5_9}.

\printbibliography

@dataset{3865-01,
  author       = {D'Amato, Marina and
                  Weber, Rianne and
                  Lefkes, Judith and
                  van der Graaf, Fennie and
                  Stegeman, Michelle and
                  Grisi, Clément and
                  Builtjes, Luc and
                  Philipp, Lena and
                  Bosma, Joeran S. and
                  Jacobs, Colin and
                  Huisman, Henkjan and
                  Litjens, Geert and
                  van der Laak, Jeroen and
                  Ciompi, Francesco and
                  Hering, Alessa},
  title        = {The UNICORN challenge: public few-shots},
  month        = jun,
  year         = 2025,
  publisher    = {Zenodo},
  version      = {7.0},
  doi          = {10.5281/zenodo.15680730},
  url          = {https://doi.org/10.5281/zenodo.15680730},
}

@article{3865-02,
  title={A real-world dataset and benchmark for foundation model adaptation in medical image classification},
  author={Wang, Dequan and Wang, Xiaosong and Wang, Lilong and Li, Mengzhang and Da, Qian and Liu, Xiaoqiang and Gao, Xiangyu and Shen, Jun and He, Junjun and Shen, Tian and others},
  journal={Scientific Data},
  volume={10},
  number={1},
  pages={574},
  year={2023},
  publisher={Nature Publishing Group UK London}
}

@article{3865-03,
  title={Fairmedfm: fairness benchmarking for medical imaging foundation models},
  author={Jin, Ruinan and Xu, Zikang and Zhong, Yuan and Yao, Qingsong and QI, DOU and Zhou, S Kevin and Li, Xiaoxiao},
  journal={{Advances in Neural Information Processing Systems}},
  volume={37},
  pages={111318--111357},
  year={2024}
}

@misc{3865-04,
  author = {Habib, Nathan and Fourrier, Clémentine and Kydlíček, Hynek and Wolf, Thomas and Tunstall, Lewis},
  title = {LightEval: A lightweight framework for LLM evaluation},
  year = {2023},
  version = {0.11.0},
  url = {https://github.com/huggingface/lighteval}
}

@article{3865-05,
  title={Sustainable data analysis with {Snakemake}},
  author={M{\"o}lder, Felix and Jablonski, Kim Philipp and Letcher, Brice and Hall, Michael B and van Dyken, Peter C and Tomkins-Tinch, Christopher H and Sochat, Vanessa and Forster, Jan and Vieira, Filipe G and Meesters, Christian and others},
  journal={F1000Research},
  volume={10},
  pages={33},
  year={2025}
}

@inproceedings{3865-06,
  title={Slurm: simple {Linux} utility for resource management},
  author={Yoo, Andy B and Jette, Morris A and Grondona, Mark},
  booktitle={Workshop on job scheduling strategies for parallel processing},
  pages={44--60},
  year={2003},
  organization={Springer}
}

@misc{3865-07,
  author       = {Alves, Natália and
                  Schuurmans, Megan and
                  Rutkowski, Dawid and
                  Yakar, Derya and
                  Haldorsen, Ingfrid and
                  Liedenbaum, Marjolein and
                  Molven, Anders and
                  Vendittelli, Pierpaolo and
                  Litjens, Geert and
                  Hermans, John and
                  Huisman, Henkjan},
  title        = {The {PANORAMA} study protocol: pancreatic cancer diagnosis -- radiologists meet {AI}},
  month        = jan,
  year         = 2024,
  publisher    = {Zenodo},
  doi          = {10.5281/zenodo.10599559},
  url          = {https://doi.org/10.5281/zenodo.10599559},
}

@article{3865-08,
  title={Artificial intelligence and radiologists in prostate cancer detection on MRI (PI-CAI): an international, paired, non-inferiority, confirmatory study},
  author={Saha, Anindo and Bosma, Joeran S and Twilt, Jasper J and van Ginneken, Bram and Bjartell, Anders and Padhani, Anwar R and Bonekamp, David and Villeirs, Geert and Salomon, Georg and Giannarini, Gianluca and others},
  journal={The Lancet Oncology},
  volume={25},
  number={7},
  pages={879--887},
  year={2024},
  publisher={Elsevier}
}

@article{3865-09,
  title={Vision foundation models for computed tomography},
  author={Pai, Suraj and Hadzic, Ibrahim and Bontempi, Dennis and Bressem, Keno and Kann, Benjamin H and Fedorov, Andriy and Mak, Raymond H and Aerts, Hugo JWL},
  journal={arXiv Preprint arXiv:2501.09001},
  year={2025}
}

@article{3865-10,
  title={Curia: A multi-modal foundation model for radiology},
  author={Dancette, Corentin and Khlaut, Julien and Saporta, Antoine and Philippe, Helene and Ferreres, Elodie and Callard, Baptiste and Danielou, Th{\'e}o and Alberge, L{\'e}o and Machado, L{\'e}o and Tordjman, Daniel and others},
  journal={arXiv Preprint arXiv:2509.06830},
  year={2025}
}

@article{3865-11,
  title={Overcoming data scarcity in biomedical imaging with a foundational multi-task model},
  author={Sch{\"a}fer, Raphael and Nicke, Till and H{\"o}fener, Henning and Lange, Annkristin and Merhof, Dorit and Feuerhake, Friedrich and Schulz, Volkmar and Lotz, Johannes and Kiessling, Fabian},
  journal={Nature Computational Science},
  volume={4},
  number={7},
  pages={495--509},
  year={2024},
  publisher={Nature Publishing Group US New York}
}

@article{3865-12,
  title={{DINOv2}: learning robust visual features without supervision},
  author={Oquab, Maxime and Darcet, Timoth{\'e}e and Moutakanni, Th{\'e}o and Vo, Huy and Szafraniec, Marc and Khalidov, Vasil and Fernandez, Pierre and Haziza, Daniel and Massa, Francisco and El-Nouby, Alaaeldin and others},
  journal={arXiv Preprint arXiv:2304.07193},
  year={2023}
}

@article{3865-13,
  title={Dinov3},
  author={Sim{\'e}oni, Oriane and Vo, Huy V and Seitzer, Maximilian and Baldassarre, Federico and Oquab, Maxime and Jose, Cijo and Khalidov, Vasil and Szafraniec, Marc and Yi, Seungeun and Ramamonjisoa, Micha{\"e}l and others},
  journal={arXiv Preprint arXiv:2508.10104},
  year={2025}
}

@Online{3865-15,
  accessed = {2025-10-29},
  author   = {{NVIDIA-NeMo}},
  title    = {{NeMo} evaluator {SDK}},
  url      = {https://github.com/NVIDIA-NeMo/Evaluator/},
}

\end{document}